\documentclass[conference]{IEEEtran}
\IEEEoverridecommandlockouts

\usepackage[sort,numbers]{natbib}
\usepackage{amsmath,amssymb,amsfonts}
\usepackage{algorithmic}
\usepackage{graphicx}
\usepackage{textcomp}
\usepackage{xcolor}
\usepackage{booktabs}
\usepackage{mathtools}
\usepackage{multirow}

\def\BibTeX{{\rm B\kern-.05em{\sc i\kern-.025em b}\kern-.08em
    T\kern-.1667em\lower.7ex\hbox{E}\kern-.125emX}}
\begin{document}

\title{Informed Decision-Making through Advancements in Open Set Recognition and Unknown Sample Detection}

\author{\IEEEauthorblockN{1\textsuperscript{st} Atefeh Mahdavi}
 \IEEEauthorblockA{\textit{Dept. of Engineering and Sciences} \\
 \textit{Florida Institute of Technology}\\
 Melbourne, USA \\
 amahdavi@fit.edu}
 \and
 \IEEEauthorblockN{2\textsuperscript{nd} Marco Carvalho}
 \IEEEauthorblockA{\textit{Dept. of Engineering and Sciences} \\
 \textit{Florida Institute of Technology}\\
 Melbourne, USA \\
 mcarvalho@cs.fit.edu}
}

\maketitle
\IEEEpubidadjcol
\begin{abstract}
  Machine learning-based techniques open up many opportunities and improvements to derive deeper and more practical insights from data that can help businesses make informed decisions. However, the majority of these techniques focus on the conventional closed-set scenario, in which the label spaces for the training and test sets are identical. Open set recognition (OSR) aims to bring classification tasks in a situation that is more like reality, which focuses on classifying the known classes as well as handling unknown classes effectively. In such an open-set problem the gathered samples in the training set cannot encompass all the classes and the system needs to identify unknown samples at test time. On the other hand, building an accurate and comprehensive model in a real dynamic environment presents a number of obstacles, because it is prohibitively expensive to train for every possible example of unknown items, and the model may fail when tested in testbeds. This study provides an algorithm exploring a new representation of feature space to improve classification in OSR tasks. The efficacy and efficiency of business processes and decision-making can be improved by integrating OSR, which offers more precise and insightful predictions of outcomes. We demonstrate the performance of the proposed method on three established datasets. The results indicate that the proposed model outperforms the baseline methods in accuracy and F1-score.
\end{abstract}


\section{Introduction}
\begin{figure}[thb]
    \centering
	\includegraphics[trim={0.001cm 0.001cm 0.001cm 0.001cm}, clip,width=0.75\linewidth,  height=0.9\linewidth]{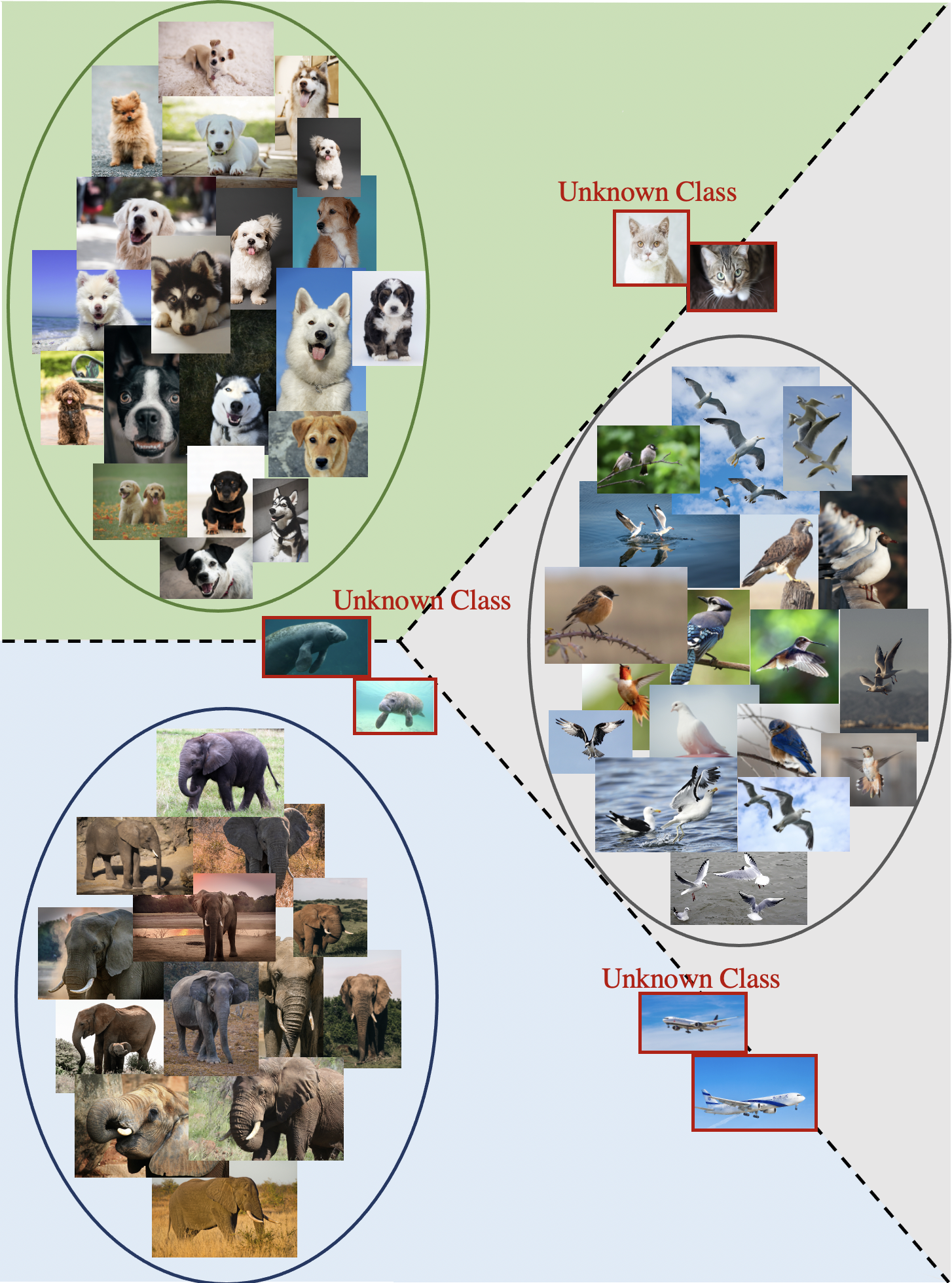}
	\caption{A closed-set classifier distinguishes known classes like dogs, birds, and elephants, but it wrongly categorizes unknown samples like cats and airplanes as known ones during testing.}
	\label{fig:closed-set}       
\end{figure}

Artificial intelligence (AI) evolves decision-making and alters its dynamics by leveraging intelligent automation that can replicate human mental processes. At present, machine learning systems are widely used in numerous commercial and industrial products, such as autonomous vehicles, video surveillance systems, manufacturing, medical imaging, and so on. These applications often involve the emergence of samples from classes that were not seen during training, commonly referred to as "unknown unknowns" (\cite{scheirer2012toward}).
Traditional training methods operate under the closed-set assumption that all classes encountered during testing are known. However, this assumption becomes problematic when the model encounters an unknown class, as it is forced to classify it as one of the known classes (see Figure \ref{fig:closed-set}). This leads to a decline in performance as the model struggles to accurately categorize the unknown class.
This limitation restricts the application of machine learning systems to known objects and hinders their functionality in real-world scenarios. 
Just like humans have the capacity to adapt, learn, and make decisions when faced with unfamiliar situations that go beyond their existing knowledge, the dynamic process of decision-making transformation enabled by intelligent automation also necessitates the ability to handle unknown elements. However, constructing a comprehensive and effective model in a dynamic environment poses challenges, as it is impractical to collect, label, and train the model on every possible instance of an unknown object. 

OSR task involves two objectives: classifying known classes and rejecting unknown classes (\cite{mahdavi2021survey}). By integrating these goals, OSR enables the development of a more robust system compared to traditional classifiers. This system establishes a more realistic environment and brings benefits to a range of applications such as self-driving cars, e-commerce product classification, video surveillance, classification and malware detection. In safety-critical applications like autonomous driving or medical diagnosis systems, OSR can aid in identifying anomalies or outliers in the data, which can offer insightful information for making decisions. For instance, consider a model trained to identify medical images. When encountering an unknown image that it cannot accurately classify, this could indicate the presence of an uncommon or atypical pathology that requires human intervention for diagnosis confirmation and further examination. \par
In this paper, we focus on representation learning, also known as feature learning, for OSR. Representation learning refers to the process of acquiring data representations that facilitate the extraction of meaningful information. It enables a machine to automatically find the representations required for detection or classification after being fed with raw data. Deep Neural Networks (DNNs) would be used to provide a set of representative features driven by various goals. 
By employing loss functions such as cross-entropy, the hidden layer representations are trained to reduce classification loss of the output in closed-set classification scenarios. However, it is important to note that the representations may not inherently contain the desired feature space that proves useful for OSR. We propose a mechanism to enhance a neural network's feature space representation for better detection of unknown situations and handling OSR. This strategy is built on expanding the existing loss functions with a new type of loss that we refer to as "Superlative Loss". Our contributions include the following:
\begin{itemize}
    \item The proposed method is effective for OSR problems in DNNs and is flexible to be used with different types of loss functions on any neural architecture.

    \item We introduce superlative loss for developing an ideal feature space representation that makes OSR easier without borrowing additional data or generating them. Thus, this robust model does not require complex network architectures which can be costly and time-consuming.

    \item We demonstrate that superlative loss delivers statistically significant improvements in terms of overall F1 score and accuracy when applied to two different types of loss functions on three datasets.
\end{itemize}

The remainder of this paper is structured as follows. We provide a summary of related works in Section 2, and in Section 3 we outline our methodology. In Section 4, we describe our evaluation methodology, findings, and additional remarks. 

\section{Literature Review}
The issue of open set recognition (OSR) was first observed in the field of face recognition and later found to be widespread in various other domains. From a modeling perspective, existing OSR techniques are categorized into discriminative models and generative models  (\cite{geng2020recent}). Discriminative models, in turn, can be classified into traditional machine learning-based methods and deep learning-based methods. Various efforts have been made to adapt traditional methods to the OSR scenario. The "1-vs-set SVM" method (\cite{scheirer2012toward}) was initially formalized OSR. This approach restricts the decision space of each class by adding an extra hyperplane, effectively rejecting unknown classes. Subsequently, the Weibull calibrated SVM (W-SVM) (\cite{scheirer2014probability}) and PI-SVM (\cite{jain2014multi}) tried to calibrate the confidence scores of different classes with statistical extreme value theory.
(\cite{bendale2015towards}) extended the nearest class mean classifier (\cite{mensink2013distance}), and performs classification by utilizing a distance calculation between unknown and known class centers. In a related context, (\cite{herbei2006classification}) introduces a binary classification that allows for a reject option in which case no decision is made. This option becomes particularly useful for instances where conditional class probabilities are in close proximity, posing challenges in their classification. These earlier techniques heavily relied on manually designed features, and their accuracy was greatly influenced by the choice of feature descriptors. In the field of related subjects, metric learning is notable for its connection with representation learning and the development of new ways to measure similarities between objects. Learning such a new metric in a traditional machine learning setting is known to enhance the performance of classification. However, potential benefits may arise from integrating metric learning with open set recognition methods. 
Recent approaches have shifted their research focus towards utilizing deep learning for detecting unknowns, benefiting from its strengths in feature extraction and automatic learning. DNNs have the advantage of learning more accurate feature representations since the features are learned simultaneously with the classification stage in an end-to-end manner. However, even the most advanced DNNs often overfit the training data and make high-confidence predictions when faced with an unknown instance. This means that DNNs are unable to identify unknown objects unless combined with an additional method or process. The OpenMax method, proposed by (\cite{bendale2016towards}), was the first to leverage deep learning for solving OSR problems. Its objective was to overcome the limitations of the Softmax function in DNNs. This was achieved by replacing the Softmax layer with a Weibull distribution fitting score, which enabled the computation of pseudo-activation for unknown classes. 
OpenMax does not inherently enhance feature representation to improve unknown detection. The difficulty in employing the distance from the Mean Activation Vector (MAV) within OpenMax lies in the fact that standard loss functions, like cross entropy, do not lead to the direct projection of class instances around the MAV.  Moreover, the difference between the testing distance function and the one used during training raises concerns about the suitability of the distance metric for the specific space. To address these concerns, our method introduces an innovative instance representation that aims to overcome these limitations. Another advanced approach, known as CROSR (\cite{yoshihashi2019classification}), builds upon insights from OpenMax's formulation. This technique leverages reconstructive latent representations to encode more about input. Their method involves enhancing a deep open-set classifier with latent representation learning for reconstruction purposes. 
In a related study (\cite{shu2017doc}), the Softmax layer was substituted with a one-vs-rest final layer of sigmoids. These methods rely on setting a threshold to distinguish between known and unknown samples, which presents the difficulty of threshold calibration. Other research works try to generate additional training data with generative models. Several methods, such as the one proposed by (\cite{ditria2020opengan}), utilize GANs as the foundation for data generation. Among these techniques, G-OpenMax (\cite{ge2017generative}), an improved version of OpenMax, takes a generative approach by creating synthetic instances of unknown classes using known class data. These synthetic samples are then used to train DNNs, enhancing the classifier's ability to identify unknown classes. Another technique presented by (\cite{neal2018open}) generates artificial open set examples known as counterfactual samples using an encoder-decoder GAN architecture. 
An alternative approach, described by (\cite{chen2021adversarial}), involves reciprocal points and adversarial learning. Reciprocal points, unlike counterfactual samples, are notably dissimilar to the known class samples. Although open set recognition techniques, often leveraging GANs to generate additional training data, have demonstrated impressive performance in addressing the open set challenge, they do have certain limitations to consider. The effectiveness of generative approaches relies on the reliability and validity of the generated data. It is unlikely that the additional data will accurately approximate the diverse real-world unknown test samples encountered during the testing stage, especially when the generated unknown class samples resemble the known class samples in appearance. Furthermore, these techniques necessitate more complex network architectures, which consequently contribute to increased computational complexity. While this added complexity can capture intricate patterns, it also introduces computational overhead. In contrast, the open set recognition technique presented in this paper stands out by not requiring the generation of additional data. This approach avoids the complexities associated with generative methods while still delivering accurate results.

\section{Approach}
As the hidden layers of DNNs can be viewed as multiple levels of input representations, the superlative loss procedure revolves around determining how to replace the original feature representation with one that is more advantageous for OSR. This new representation can be learned in such a way that different classes are further apart and well separated which leads to larger spaces among them. Consequently, unknown examples can easily be detected. Suppose $K$ represent the total number of distinct known classes, and let $D={(x_{1}, y_{1}),(x_{2}, y_{2}),...,(x_{N}, y_{N})}$ denote a training dataset consisting of $N$ samples $x_{i} \in X = {x_{1},x_{2},...,x_{N}}$ and their corresponding labels $y_{i} \in Y = {1,2,...,K}$. OSR involves learning a function $f$ that can accurately classify an instance into one of the known classes or an unknown class. For an unseen instance $x^{'}$ (not present in $X$), if $y^{'}=f(x^{'})$ represents the class label predicted by $f$, $y^{'}$ can potentially belong to one of the recognized $K$ classes in a closed set scenario, or it can indicate a new class in an open set scenario.
In the proposed algorithm the between-class separation is maximized in terms of the distance between class means of all the $K$ known classes. Then, during neural network training, we attempt to declare the desired space using the original space. 

\subsection{Superlative Loss Function}
\begin{figure*}[h]
    \centering
	\includegraphics[trim={0.001cm 0.001cm 0.001cm 0.001cm}, clip,width=0.7\linewidth, height=0.45\linewidth]{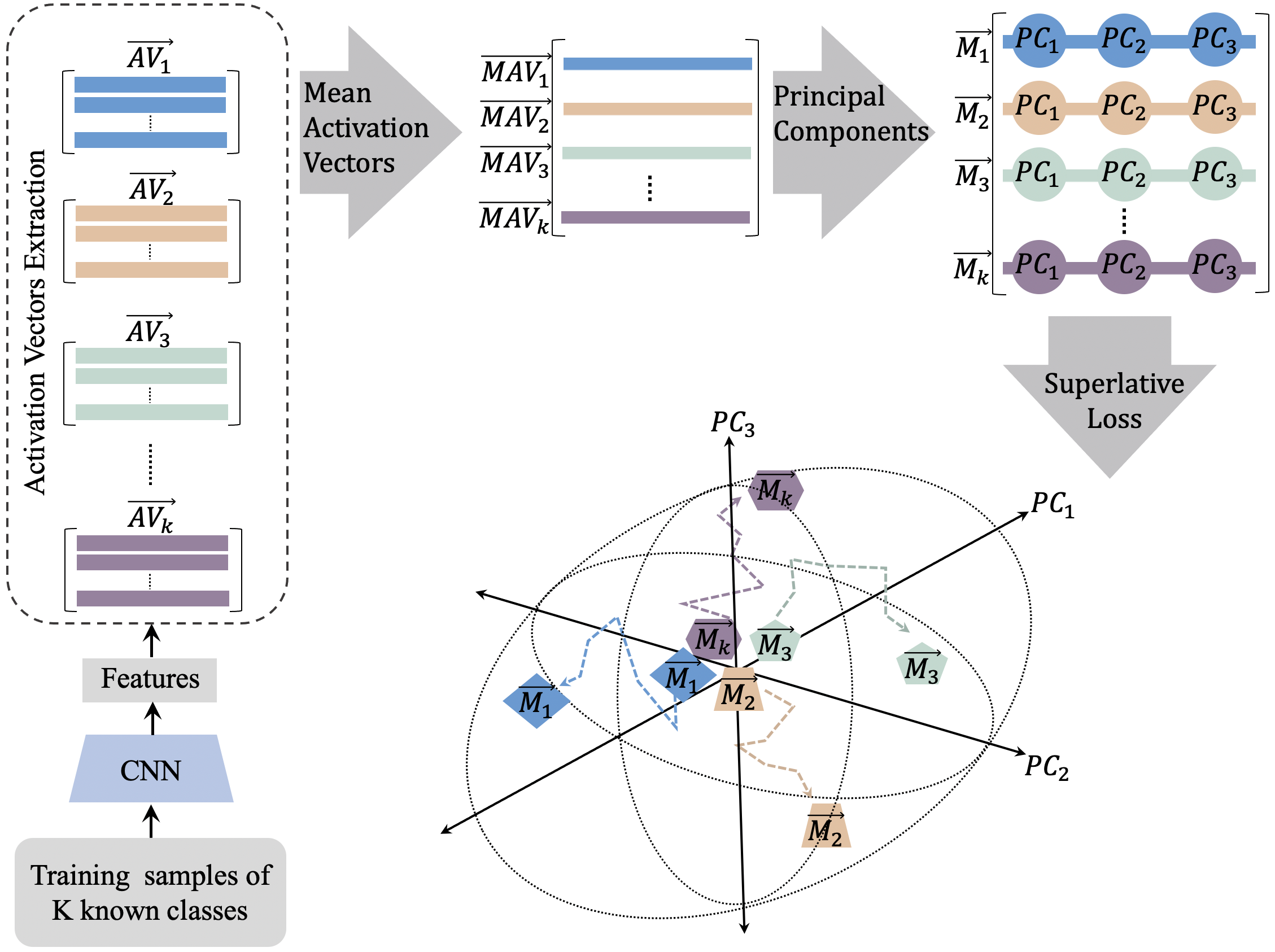}
	\caption{An overview of proposed method.}
	\label{fig:methodology}       
\end{figure*}

In this study, our focus is on the network values obtained from the penultimate layer, which corresponds to the fully connected layer preceding the SoftMax function. These values are responsible for extracting higher-level representations from the input data. We refer to these values as the activation vector ($\smash{\overrightarrow{AV}}$).
For open set images, the activation vector typically exhibits a small magnitude. This small magnitude can be attributed to the absence of the unknown class during the training phase, which hinders the network's ability to learn its distinctive features. By leveraging the characteristics of this layer, we incorporate the information derived from the $\smash{\overrightarrow{AV}}$ into our approach. The training phase of the proposed algorithm is depicted in Figure \ref{fig:methodology}. In this figure, the activation vectors ($\smash{\overrightarrow{AV}}$) of each training sample belonging to the $k$ known classes are visualized, with distinct colors assigned to each of the $k$ known classes. The algorithm begins by computing the means of the known classes as an initial step. To accomplish this, the mean activation vectors ($\smash{\overrightarrow{MAV}}$) for each class are calculated by averaging the $\smash{\overrightarrow{AV}}$ values of the training instances associated with that class:

\begin{align}
\smash{\overrightarrow{MAV}_{i}}
= \underset{\text{$1 \le i \le K$}} {\frac{1}{N_{i}}\sum_{n= 1}^{N_{i}} \smash{\overrightarrow{AV}_{i,n}}} 
\end{align}

In the above equation, for the $K$ known classes, $N_{i} $ is the number of training examples in each class. We utilize principal component analysis (PCA) to extract the principal components from the mean activation vectors of each class. Such approaches (\cite{wang2019backpropagation}) enables us to efficiently reduce the dimensionality of the data while preserving essential information encapsulated in these components. The three highest-ranking principal components, denoted as $PC_{1}$, $PC_{2}$, and $PC_{3}$, are selected. Employing a restricted number of principal components not only reduces the number of variables in the optimization process of declaring superlative space, leading to less time consumption but also ensures that all features have the desired impact based on their significant portion of the total variance. Additionally, it enhances data exploration and visualization efficiency through dimensionality reduction, while also tackling the curse of dimensionality issue in high-dimensional spaces. This leads to more robust distance calculations that remain resilient against noise distortion. Consequently, the features are represented in the feature space by three coordinates. In our approach, each class is represented by a single point, denoted as $\smash{\overrightarrow{M}}$, which captures its projection onto the three selected principal components.
After completing all the defined steps and variables, our main objective is to maximize the distance between the represented points ($\smash{\overrightarrow{M}}s$) within the feature representation. To achieve this, we establish a boundary that encompasses the points. To determine the radius of this boundary, we prioritize the first principal component, as it contributes the most to the overall variation. By examining the maximum internal distance between the minimum and maximum values of $PC_{1}$ vectors across all training samples within the dataset, we obtain the maximum spread length. This approach guarantees that the boundary radius is confined within the bounds of the variance captured through $PC_{1}$.
The radius of the boundary defined by a coefficient, denoted as the hyperparameter $\Gamma$ ($\Gamma \in \mathbb{R} | \Gamma > 1$) in equation 2, effectively enhances the inter-class distance when combined with the first-ranked maximum internal distance.
Subsequently, our aim is to minimize the distance between each point's current position $\smash{\overrightarrow{M}_{i}}$ and the boundary, compelling them to move closer to it. This process ensures maximum separation between points. We refer to this characteristic as the "boundary distance," denoted by $\textit{BD}$:

\begin{align}
\textit{BD} = \sum_{i=1}^{K} \left[ \Gamma * \left(   \max(\smash{\overrightarrow{PC_{1}}}) - \min(\smash{\overrightarrow{PC_{1}}}) \right) - \left \| \smash{\overrightarrow{M}_{i}} \right \|_{2}^{2} \right]
\end{align}

During the experiment, we observed a situation where two points from different classes come into close proximity to each other as they approach the boundary. 
This undesired situation undermines the objective of maximizing the inter-class distances. To address this challenge, we introduce an additional constraint that ensures that the distance between each point and its nearest neighbor is maximized as they approach the boundary. This characteristic, known as "inter-separation", is denoted by $\textit{IS}$ and is defined as:  

\begin{align}
\underset{\substack{1 \leq  i \neq j \leq K}} {\textit{IS} ={\max}\left \| \smash{\overrightarrow{M}_{i}} - \smash{\overrightarrow{M}_{j}}\right \|_{2}^{2}
}
\end{align}

By imposing this constraint, the algorithm ensures that points representing different classes maintain a sufficient separation.
This helps to preserve the performance of the algorithm in terms of overall classification accuracy and maximization of inter-class distances.
To further enhance the feature space and optimize performance, we also minimize the distance of each sample to its corresponding class mean. The term "intra-compactness", denoted by $\textit{IC}$, is used to describe this characteristic and defined as:
\begin{align}
\textit{IC} = \sum_{i=1}^{K} \sum_{n=1}^{N_{i}} \left \| \smash{\overrightarrow{M}_{i}} - \smash{\overrightarrow{PC}_{n}}\right \|_{2}^{2} 
\end{align}
where $\smash{\overrightarrow{PC}_{n}}$ represents the principal components of each of the $N$ samples from the $K$ known classes. By applying this equation, we compute the Euclidean distance between each sample and its corresponding class mean.
The superlative loss, which is defined as a combination of the properties of boundary distance, inter-separation, and intra-compactness, is utilized to train the network. This loss function, denoted as $\mathcal{L}_{s}$, is minimized using mini-batch stochastic gradient descent with backpropagation, and described as: 

\begin{equation}
\mathcal{L}_{s} = \textit{BD} - \textit{IS} + \textit{IC} 
\end{equation}

\subsection{Prediction for Known and Unknown Classes}
During the inference phase of OSR, the main task is to classify a set of $K+1$ labels into $K+1$ distinct categories. Among these labels, the first $K$ labels correspond to the known classes that the classifier has been trained on. The remaining label, $(K+1){\text{st}}$, is specifically assigned to represent the unknown class, indicating that an instance does not belong to any of the recognized classes. To achieve this classification, it is essential to establish a threshold value that acts as a criterion for distinguishing between the known classes and the unknown class. We adopted an approach where we estimate a separate threshold value for each class instead of applying a uniform threshold across all classes. Through our experiments, we observed that estimating a per-class threshold provides more accurate results. By tailoring the threshold to the specific characteristics of each class, we can effectively account for variations in the data distribution across different classes. To determine the class-specific threshold, we follow a specific procedure. This threshold value defines the minimum distance required between an instance and its nearest class mean for it to be classified as an unknown. After completing the training phase and updating the network's weights, we obtain learned superlative representation ($\smash{\overrightarrow{MAV}}$) for each of the $K$ known classes. Subsequently, we calculate the distances between $\smash{\overrightarrow{AV}}$ of each training sample and its corresponding $\smash{\overrightarrow{MAV}}$. For each individual class, these distances are sorted in ascending order to capture the largest of the distances. Then, we select the distance value at the 99th percentile as the class-specific threshold.
During testing, we measure the distance between the $\smash{\overrightarrow{AV}}$ of a test sample and the class means. This distance is then compared to the threshold value defined specifically for each class.  If the distance exceeds the cutoff, it means that the sample is significantly further away from the class means. In this case, the sample is categorized as an unknown class and labeled as $K+1$. Conversely, if the distance falls below the threshold, it indicates that the test sample is relatively closer to one of the known classes. The final prediction is then determined by identifying the nearest class mean among the $K$ known classes. The test sample is assigned the label corresponding to the class with the closest mean with the highest probability $P$:
\begin{equation}
    y =
    \begin{dcases*}
        K+1, & \text{if distance $>$ threshold} \\
        \underset{1 \le i \le K}{\text{argmax}}\; P(y = i \;|\; \overrightarrow{x}), & \text{otherwise}
    \end{dcases*}
\end{equation}

It is worth mentioning that, the potential limitations of the proposed $\smash{\overrightarrow{MAV}}$ may exist  when dealing with multi-modal data probability distributions within a single label. Challenges arise due to the requirement of accurately representing diverse modes within the distribution of a single label using only a single mean vector. To address these challenges, future research could explore the incorporation of alternative techniques such as Variational Autoencoders, and Generative Adversarial Networks. These techniques can effectively capture the complexities of multi-modal distributions through the modeling of distinct modes. The investigation and integration of these approaches in future studies could enhance the adaptability of the proposed methodology to diverse data scenarios and contribute to a more comprehensive understanding of its applicability.

\section{Experimental Analysis}

First, this section will provide an introduction to the specifics of the dataset and the evaluation schemes. Next, we will outline the experimental setup and delve into the exploration analysis.
\subsection{Implementation Details}
\begin{figure*}[thb]
    \centering
\includegraphics[width=0.75\linewidth, height=0.2\linewidth]{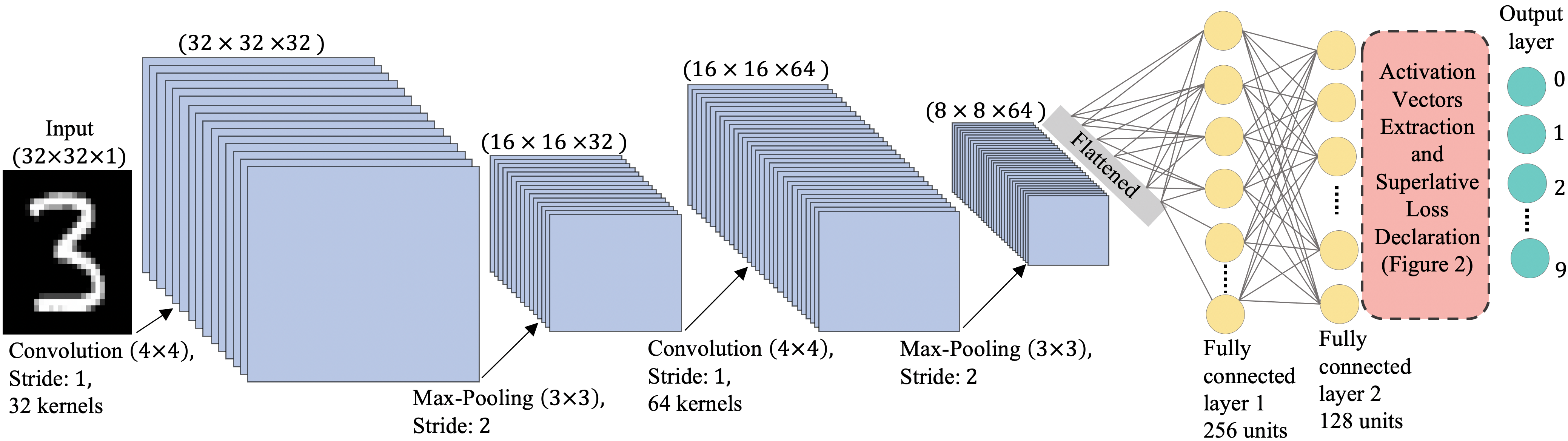}
	\caption{Visualization of the network architecture.}
	\label{fig: Architecture}      
\end{figure*}

\begin{figure*}
    \centering
	\includegraphics[trim={0.001cm 0.001cm 0.001cm 0.001cm}, clip, width=0.9\linewidth]{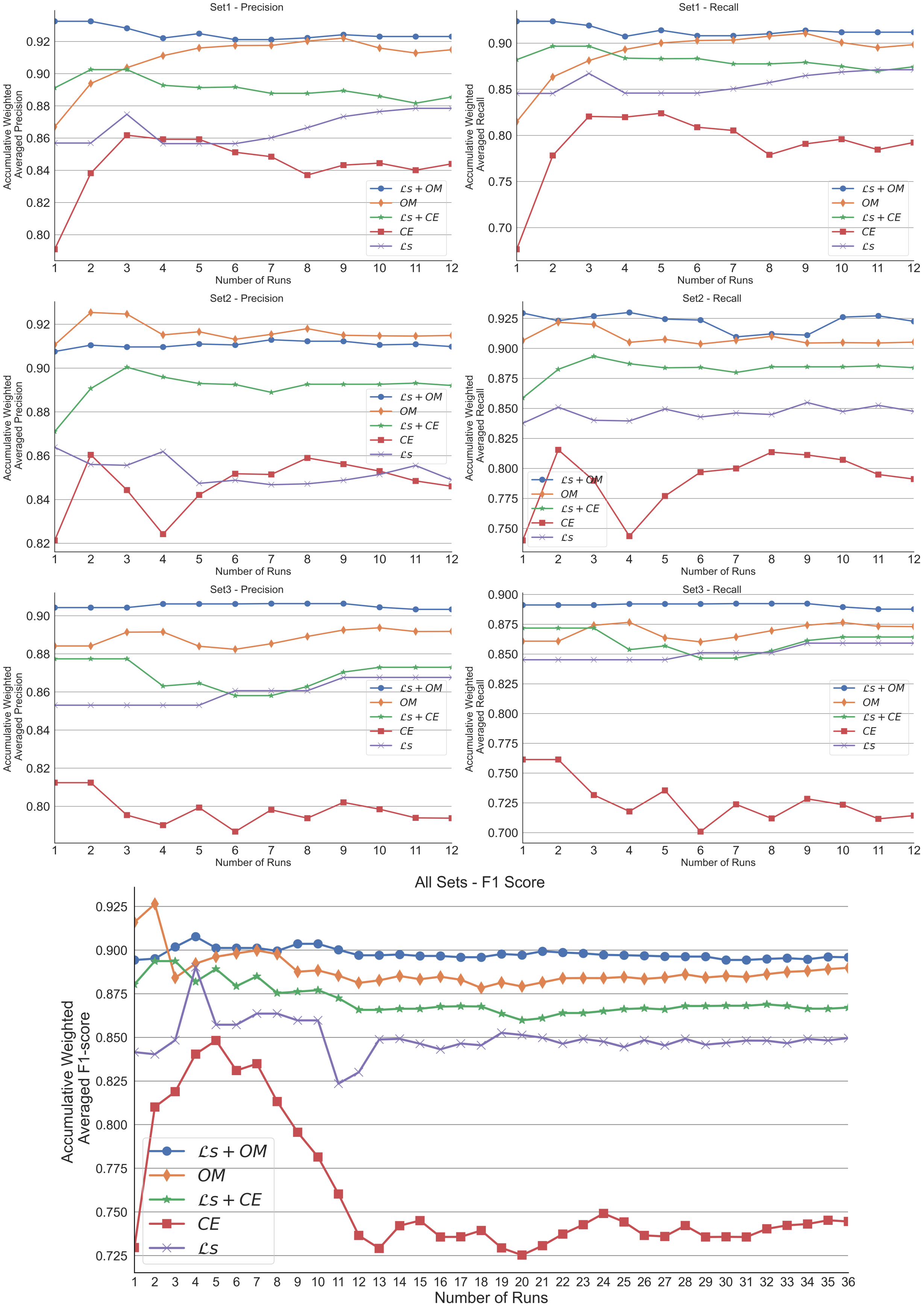}
	\caption{This figure compares precision, recall, and F1 scores for MNIST test samples from three sets ($\textbf{Set{1}}$, $\textbf{Set{2}}$, and $\textbf{Set{3}}$) across 12 runs, considering all methods. It also displays the accumulative F1 score over 36 runs, combining results from all sets.}
	\label{fig:total_test}       
\end{figure*}

Figure \ref{fig: Architecture} represents the network architecture implemented for this experiment using TensorFlow, which consists of convolutional layers, max-pooling layers, and fully connected layers. The specific configuration of these layers can be observed in the provided diagram. After the convolutional layers, two fully connected non-linear layers are employed. The output of the last layer undergoes a Softmax function, resulting in the generation of a probability distribution across the known classes. We employ the Rectified Linear Unit (ReLU) activation function for all non-linear layers and set the keep probability of Dropout to 0.2 for the fully connected layers. Batch normalization is applied to all layers. To train our networks, we utilize the Adam optimizer with a learning rate of 0.001 for a total of 3000 iterations.
We assessed the performance of our approach using a set of three distinct datasets.\\
\textbf{MNIST} consists of 70,000 gray scale images of handwritten numbers from 0 to 9. For each class, around 6,000 training samples and 1000 test samples are used.\\ 
\textbf{Fashion-MNIST} is a collection of gray scale images featuring 10 classes of clothing items. It comprises 60,000 training examples and 10,000 testing examples.\\ 
\textbf{CIFAR-10} includes 60,000 32x32 color images distributed across 10 distinct classes, with each class containing 6,000 images. The dataset is divided into 50,000 training images and 10,000 test images. In our experimental setup, we transform the color images into gray scale.\\
In each dataset, we randomly select six classes as the known classes during the training phase, while the remaining classes are considered unknown during testing. As a result, we remove the instances belonging to unknown classes from the training set. However, the test set remains unchanged, containing both known and unknown class instances. This methodology enables us to simulate open-set recognition scenarios across all datasets. For each dataset, we create three distinct groups of open-set datasets referred to as $Set{1}$, $Set{2}$, and $Set{3}$. Each set consists of a random selection of six known classes, while the remaining four classes are designated as unknown classes. For example, the specific sets chosen for MNIST dataset for evaluation in this study are defined as $Set{1} = \left [ 0, 2, 3, 4, 6, 9 \right ]$, $Set{2} = \left [ 0, 1, 2, 5, 7, 8 \right ]$, and $Set{3} = \left [ 0, 1, 3, 4, 7, 8 \right ]$. All other digits are considered as the unknown class in each set, respectively. To assess the performance, we conduct a total of 36 runs, with 12 experiments conducted for each of $Set{1}$, $Set{2}$, and $Set{3}$. We determine the specific number of runs based on our objective to find the optimal number that consistently produces stable performance. To achieve this, we analyze the relationship between the number of runs and performance metrics. The findings, depicted in Figure \ref{fig:total_test}, reveal that performance becomes stable after 12 runs for all the methods employed.\par
We conducted an evaluation of five different approaches as our implementation framework. The first approach involves training a network, as shown in Figure \ref{fig: Architecture}, using the superlative loss ($\mathcal{L}_{s}$).
As a baseline, the second approach focused on training a network using only cross-entropy loss ($CE$). The third approach is based on OpenMax ($OM$) (\cite{bendale2015towards}), which we re-implemented using the original paper and the authors' source code.
Considering that the superlative loss aims to enhance feature representation, it can be effectively combined with various other loss functions. In our study, we employed a combination of the superlative loss with both cross-entropy loss ($\mathcal{L}{s} + CE$) and OpenMax ($\mathcal{L}{s}$ + $OM$) as fourth and fifth approaches, respectively. 
This setup involved training the network using the $\mathcal{L}{s}$ in conjunction with $CE$ or $OM$. In the training process, the first 1500 iterations update the network weights to minimize $\mathcal{L}{s}$, and subsequently, the remaining training process is set to minimize the other respective losses.\par 
\subsection{Results and Analysis}

\begin{figure}[h]
    \centering
\includegraphics[width=0.9\linewidth]{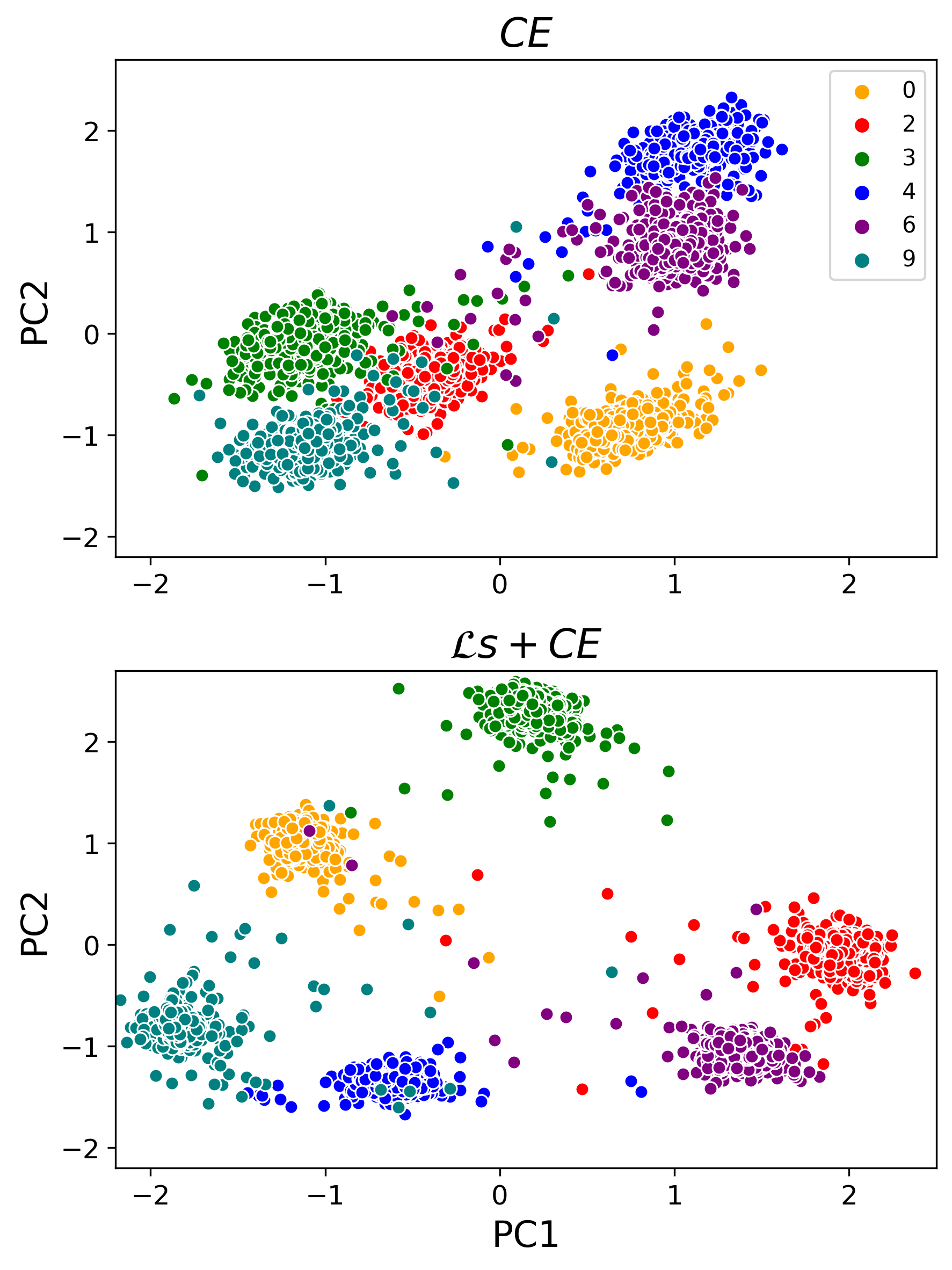}
	\caption{Feature space visualization of MNIST dataset in the experiments of $CE$ vs $\mathcal{L}{s} + CE$. Labels 0,2,3,4,6,9 represent the known classes.}
	\label{fig: PCA}      
\end{figure}

The top three rows of Figure \ref{fig:total_test} illustrate the precision and recall evaluation metrics for the test samples of MNIST dataset in $Set{1}$, $Set{2}$, and $Set{3}$, respectively. The bottom row represents the F1 score, which considers all three sets combined. The novel $\mathcal{L}{s}$ approach demonstrates an approximate performance of 85\% across all sets for all evaluation metrics. Furthermore, combining the $\mathcal{L}{s}$ approach with either the $CE$ or $OM$ methods significantly enhances their performance, particularly noticeable in the case of $\mathcal{L}{s} + CE$. Overall, in the evaluation of F1 score performance, the analysis from the bottom row of the graph reveals that the $\mathcal{L}{s} + OM$ method demonstrates superior results, achieving an approximate F1 score of 90 percent. Following that, the $OM$, $\mathcal{L}{s} + CE$, $\mathcal{L}{s}$, and $CE$ methods exhibit successively decreasing performance in that order. Table \ref{table:table1} presents the evaluation results for the case of four unknown classes out of ten classes. The average recalls, precisions, F1 scores, and accuracies are calculated for the $K$ known classes and the unknown class and then averaged across the $K + 1$ classes to obtain the Overall values. As it can be seen, $\mathcal{L}{s} + CE$ and $\mathcal{L}{s} + OM$ outperform their standalone versions. Specifically, combining $CE$ with $\mathcal{L}{s}$ results in a significant improvement across all metrics such as F1 score and accuracy. Similarly, combining $\mathcal{L}{s}$ with $OM$ leads to notable improvement in known classes for the overall F1 score and accuracy. Importantly, the $\mathcal{L}{s}$ demonstrates enhanced robustness in detecting unknown classes compared to standalone $CE$, with a notable improvement of 8.9\% in recall for unknown classes. This improvement underscores the effectiveness of the proposed method in detecting unknown classes while reducing the classification error of the training data, thereby achieving competitive results compared to standalone $CE$ in the classification task. This can be attributed to two key factors. Firstly, in the superlative representation, the learned activation vectors are more discernible compared to conventional neural network features. Secondly, the superlative loss effectively guides the feature training process, enhancing the intra-class compactness and inter-class separation of the feature representation.
\begin{table*}
  \centering
  \caption{\label{font-table} Comparison of average precisions, recalls, F1 scores, and accuracies across 36 runs.}
  \label{tab:fonts}
  \fontsize{8}{9}\selectfont
  \begin{tabular}{c|cp{0.6cm}p{0.8cm}p{0.6cm}p{0.6cm}p{0.8cm}p{0.6cm}p{0.6cm}p{0.8cm}p{0.6cm}p{0.6cm}p{0.8cm}p{0.6cm}}
    \toprule
    \multirow{2}{*}{\rotatebox[origin=c]{90}{\shortstack{\fontsize{7}{7}\selectfont Datasets}}} & \multirow{2}{*}{\fontsize{9}{8}\selectfont Methods} & \multicolumn{3}{c}{\fontsize{9}{8}\selectfont Precision} & \multicolumn{3}{c}{\fontsize{9}{8}\selectfont Recall} & \multicolumn{3}{c}{\fontsize{9}{8}\selectfont F1 Score} & \multicolumn{3}{c}{\fontsize{9}{8}\selectfont Accuracy} \\
    \cmidrule(lr){3-5} \cmidrule(lr){6-8} \cmidrule(lr){9-11} \cmidrule(lr){12-14}
    & & \fontsize{8}{8}\selectfont Overall & \fontsize{8}{8}\selectfont Unknown & \fontsize{8}{8}\selectfont Known & \fontsize{8}{8}\selectfont Overall & \fontsize{8}{8}\selectfont Unknown & \fontsize{8}{8}\selectfont Known & \fontsize{8}{8}\selectfont Overall & \fontsize{8}{8}\selectfont Unknown & \fontsize{8}{8}\selectfont Known & \fontsize{8}{8}\selectfont Overall & \fontsize{8}{8}\selectfont Unknown & \fontsize{8}{8}\selectfont Known \\
    \midrule
    \multirow{5}{*}{\rotatebox[origin=c]{90}{\shortstack{\fontsize{7}{7}\selectfont MNIST}}} & $CE$ & 0.826 & 0.823 & 0.881 & 0.761 & 0.761 & 0.84 & 0.745 & 0.750 & 0.832 & 0.761 & 0.761 & 0.840 \\
    & $\mathcal{L}{s} + CE$ & 0.880 & 0.882 & 0.928 & 0.868 & 0.875 & 0.924 & 0.867 & 0.875 & 0.924 & 0.868 & 0.875 & 0.924 \\
    & $OM$ & 0.907 & 0.905 & 0.936 & 0.892 & 0.892 & 0.927 & 0.890 & 0.889 & 0.925 & 0.892 & 0.892 & 0.927 \\
    & $\mathcal{L}{s} + OM$ & 0.912 & 0.911 & 0.981 & 0.897 & 0.898 & 0.979 & 0.896 & 0.897 & 0.979 & 0.897 & 0.898 & 0.979 \\
    & $\mathcal{L}{s}$ & 0.823 & 0.831 & 0.851 & 0.821 & 0.829 & 0.848 & 0.833 & 0.838 & 0.857 & 0.837 & 0.842 & 0.860 \\
    \midrule
    \multirow{5}{*}{\rotatebox[origin=c]{90}{\shortstack{\fontsize{7}{7}\selectfont Cifar}}} & $CE$ & 0.544 & 0.582 & 0.769 & 0.473 & 0.503 & 0.767 & 0.428 & 0.462 & 0.766 & 0.473 & 0.503 & 0.767 \\
    & $\mathcal{L}{s} + CE$ & 0.557 & 0.585 & 0.822 & 0.498 & 0.52 & 0.822 & 0.474 & 0.500 & 0.82 & 0.498 & 0.520 & 0.822 \\
    & $OM$ & 0.584 & 0.624 & 0.772 & 0.574 & 0.612 & 0.769 & 0.573 & 0.614 & 0.769 & 0.574 & 0.612 & 0.769 \\
    & $\mathcal{L}{s} + OM$ & 0.594 & 0.629 & 0.813 & 0.59 & 0.631 & 0.813 & 0.589 & 0.629 & 0.811 & 0.59 & 0.631 & 0.813 \\
    & $\mathcal{L}{s}$ & 0.553 & 0.575 & 0.806 & 0.531 & 0.562 & 0.806 & 0.535 & 0.565 & 0.804 & 0.531 & 0.562 & 0.806  \\
    \midrule
    \multirow{5}{*}{\rotatebox[origin=c]{90}{\shortstack{\fontsize{7}{7}\selectfont Fashion-\fontsize{7}{7}\selectfont MNIST}}} & $CE$ & 0.522 & 0.531 & 0.961 & 0.588 & 0.590 & 0.961 & 0.510 & 0.518 & 0.961 & 0.588 & 0.59 & 0.961 \\
    & $\mathcal{L}{s} + CE$ & 0.643 & 0.650 & 0.973 & 0.617 & 0.619 & 0.972 & 0.597 & 0.601 & 0.973 & 0.617 & 0.619 & 0.972 \\
    & $OM$ & 0.608 & 0.616 & 0.962 & 0.559 & 0.567 & 0.961 & 0.508 & 0.522 & 0.961 & 0.559 & 0.567 & 0.961 \\
    & $\mathcal{L}{s} + OM$ & 0.727 & 0.740 & 0.972 & 0.673 & 0.681 & 0.971 & 0.659 & 0.672 & 0.971 & 0.673 & 0.681 & 0.971 \\
    & $\mathcal{L}{s}$ & 0.612 & 0.625 & 0.973 & 0.611 & 0.619 & 0.972 & 0.601 & 0.616 & 0.972 & 0.611 & 0.619 & 0.972 \\
    \bottomrule
  \end{tabular}
  \label{table:table1}
\end{table*} Consequently, the combination of highly discriminative features and per-class thresholds contributes to a substantial enhancement in unknown detection performance. Figure \ref{fig: PCA} visually presents the superlative space of $Set1$ of MNIST dataset when subjected to the $CE$ and $\mathcal{L}{s} + CE$ models in a 2D space. The x-axis corresponds to the first principal component ($PC_{1}$), while the y-axis represents the second principal component ($PC_{2}$). 
From the graph, it can be observed that the six classes associated with the CE model are roughly situated in the middle of the space. In contrast, the classes of the $\mathcal{L}{s} + CE$ model are positioned around them, exhibiting greater distances between the classes and greater compactness within each class individually. This transformation serves the purpose of the $\mathcal{L}{s} + CE$ approach, which aims to create more separation between the classes, benefiting closed-set classification and improving pen set recognition. We also investigate the model's performance under different degrees of openness (\cite{scheirer2012toward}), which is defined by considering the number of classes seen during training ($C_{\text{train}}$), the number of classes in the test set ($C_{\text{test}}$), and the number of classes to be identified during testing ($C_{\text{target}}$):
\begin{align}
\text{openness} = 1 - \sqrt{\frac{2 \times C_{\text{train}}}{C_{\text{test}} + C_{\text{target}}}}
\end{align}

\begin{figure}
    \centering
	\includegraphics[width=1\linewidth]{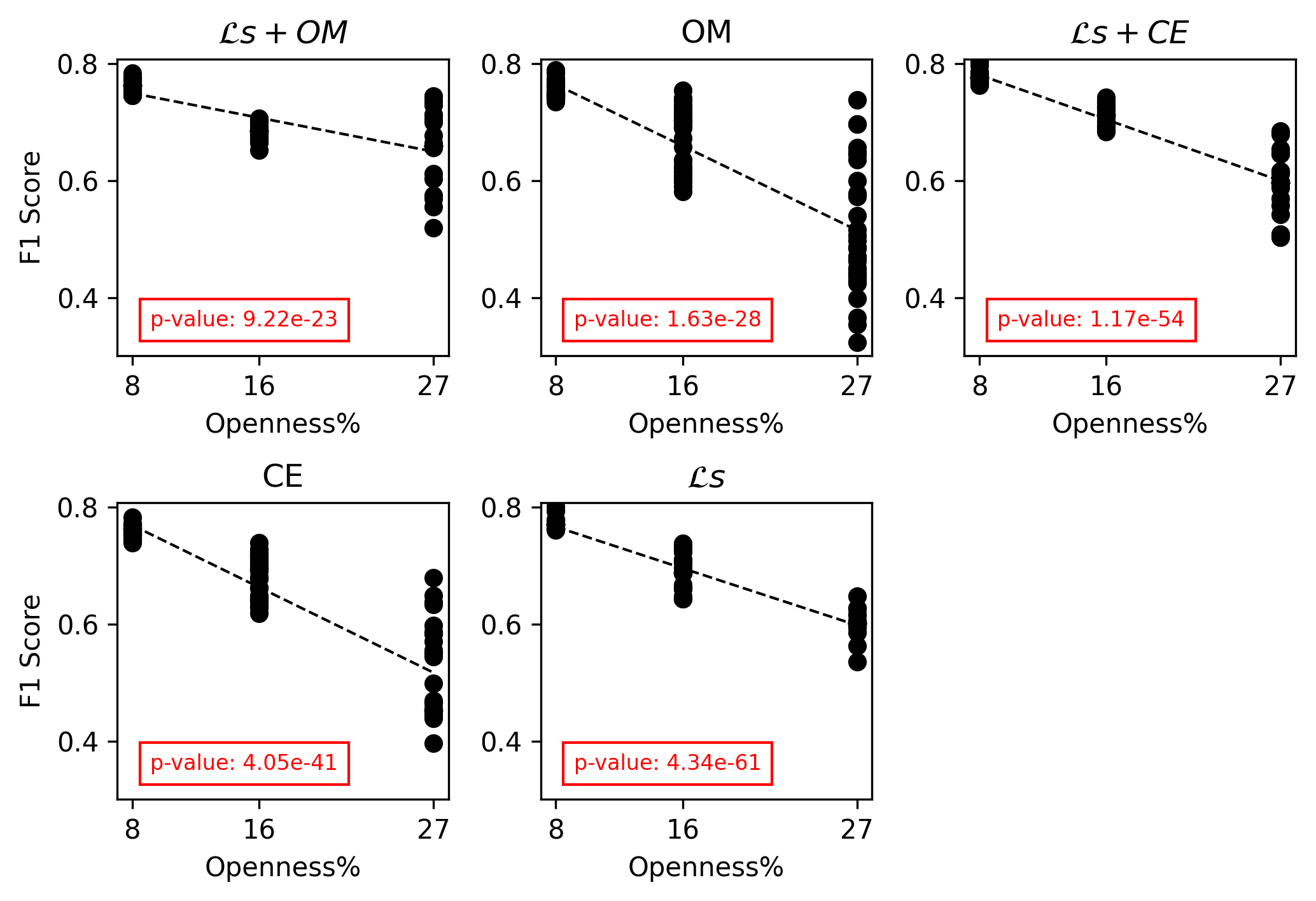}
	\caption{F1 scores against varying openness for Fashion-MNIST.}
	\label{fig: openness}      
\end{figure}

In our experimentation using the Fashion-MNIST dataset, we maintain all ten classes during the testing phase ($C_{\text{test}} = 10$). The number of known classes in the training phase varies as 8, 6, and 4, while the remaining classes are treated as the unknown class to be recognized alongside the known classes during inference ($C_{\text{target}} = C_{\text{train}} + 1 $). This setup results in openness variations of 8\%, 16\%, and 27\%. A higher openness score indicates a greater number of classes considered as unknown. The evaluation involves assessing the overall f1 scores of different models, and the results are depicted in Figure \ref{fig: openness}. This figure illustrates how the f1-score changes across different degrees of openness for each individual model. We observe that as openness increases, the overall performance of all models decreases. The rate of this decline is minimal for $\mathcal{L}{s}$ and $\mathcal{L}{s} + OM$ methods that shows more stability across values of openness for these models. We have shown the statistical significance by p\_value study.

\section{Conclusion}
The presented approach introduces a novel solution to tackle open set recognition. This method aims to reposition and enhance the compactness of features to create greater separation between samples from different classes while bringing samples from the same classes closer together. The objective is to increase the discriminative space and improve the detection of unknown samples.
The incorporation of Principal Component Analysis in the optimization process proves to be highly advantageous in terms of simulation time, visualization, and performance. Particularly, when dealing with datasets containing a large number of features, the benefits of PCA become more apparent. For future work, combining the presented method with other loss functions could be explored to investigate its impact and potential enhancements. 
Additionally, improving threshold estimation is a potential avenue for future research.

\bibliographystyle{IEEEtranN}
\bibliography{bibtex}
\end{document}